# Sensor fusion using EMG and vision for hand gesture classification in mobile applications

Enea Ceolini*, Gemma Taverni*, Lyes Khacef †, Melika Payvand* and Elisa Donati*
*Institute of Neuroinformatics, University of Zurich and ETH Zurich, Switzerland
†Universite´ Coˆte d'Azur, CNRS, LEAT, France

*Abstract*—The discrimination of human gestures using wearable solutions is extremely important as a supporting technique for assisted living, healthcare of the elderly and neurorehabilitation. This paper presents a mobile electromyography (EMG) analysis framework to be an auxiliary component in physiotherapy sessions or as a feedback for neuroprosthesis calibration. We implemented a framework that allows the integration of multi-sensors, EMG and visual information, to perform sensor fusion and to improve the accuracy of hand gesture recognition tasks. In particular, we used an event-based camera adapted to run on the limited computational resources of mobile phones. We introduced a new publicly available dataset of sensor fusion for hand gesture recognition recorded from 10 subjects and used it to train the recognition models offline. We compare the online results of the hand gesture recognition using the fusion approach with the individual sensors with an improvement in the accuracy of 13% and 11%, for EMG and vision respectively, reaching 85%.

## I. INTRODUCTION

Biopotentials are electrical signals that are generated by physiological processes occurring within the body. EMG is the measurement of the electrical potential generated by activated motor units. When the muscle is contracted, it generates electrical potentials that can be easily measured using non-invasive sensor devices placed on the skin, surface EMG (sEMG). EMG signals are very helpful in gesture recognition for rehabilitation [1] as well as in monitoring physical and sport performance [2].

A new generation of wearable devices allows continuous EMG monitoring of signals which can be helpful for physiotherapists to use both in diagnosis and treatment contexts. With the ubiquity of smartphones in the society, a mobile application represents an easy access to continuous monitoring for personalized medicine. To meet the required needs, the application should provide real-time biofeedback with task performance to help the physiotherapists to define goals.

In this paper, we propose a feature extraction and fusion methodology to perform static hand gesture classification in a mobile application, called 'RELAX'. Sensor fusion is actually a subcategory of data fusion and it is the process of combining sensory data from multiple sensors such that to reduce the amount of uncertainty in the resulting information. In particular, we consider the complementary features extracted from a visual sensor and sEMG measurements. The visual input comes from a neuromorphic event-based camera as the Dynamic Vision Sensor (DVS) [3] or its advanced extension the Dynamic and Active Pixel Vision Sensor (DAVIS) [4]. The DVS operates at high temporal resolution and low computational power allowing a new level of performance in real-time vision. Due to the limited computational resources of a mobile platform, the DVS/DAVIS camera is an optimal solution for continuous monitoring [5]. We used a complementary fusion based on EMG and camera input, since the sensors do not directly depend on each other and can be combined in order to give more information about the hand gestures.

Standard methods in EMG processing and classification focus on feature extraction that can be then fed into a remote classifier or regression system [6]. The pre-processing for feature extraction can be applied either in time (e.g. Mean Absolute Value (MAV), Root Mean Square (RMS)) [7], in frequency (e.g. Power Spectrum Density, Fast Fourier Transform) [8] or in time-frequency (e.g. Wavelet Transform) [9] domains. More recently, a new approach for classifying EMG signals started to emerge based on the use of Spiking Neural Networks (SNNs) on neuromorphic chips [10].

Event-based cameras have already been used for gesture recognition task where a Convolutional Neural Network (CNN) plays "RoShamBo" (rock, paper, scissor) against human opponents in real-time [11]. In [5], the authors present a mobile application that uses the output of an event-based camera for gesture recognition. The aim of this work is to test a data fusion method on sEMG signals recorded by the Myo armband from the forearm and improve the classification rate of hand gestures with the introduction of visual input using the computational resources of a mobile phone. Our main contribution is the first ever development of a mobile application where the EMG and vision sensor are effectively fused to increase the accuracy of a hand gesture recognition task. Moreover, we collected a dataset which we have made publicly available. There are different ways of merging multi-sensors data: at the feature level, fusing the feature vectors, or at the classifier level, combining inferences decisions from each sensor. The second approach produces better results [12], and it is what we employed in this paper.

## II. MATERIALS AND METHODS

In this section, we present the recorded dataset used for training the recognition models and introduce our sensor system connected to an Android mobile phone.

### A. DVS and DAVIS Camera

The DVS is an event-based camera, inspired by the mammalian retina [3]. The DVS encodes visual information ef-ficiently removing redundancy. Each pixel responds asynchronously to changes in brightness with the generation of events: an ON-event when the light increases, or OFF-events when the light decreases, above a certain threshold. Only the active pixels transfer information and the static background is directly removed on hardware at the front-end. The asynchronous nature of the DVS makes the sensor low power, low latency and low-bandwidth, as the amount of data transmitted is very small, making it the best solution for a mobile app. Each DVS event can be represented as a vector composed by the pixel location in the pixel array (x and y coordinates), the timestamp (with $us$ resolution), and the polarity (ON and OFF). Fig. 1 shows a typical output of the DAVIS camera [4], an advanced version of the DVS [3] which is able to record Active Pixel Sensor (APS) "conventional" frame (Fig. 1a) and DVS events (Fig. 1b). In this way, the sensor gives the possibility to benefit from the two different kinds of output modalities depending on the application.

### B. Dataset description

The dataset contains muscle activity and video recordings. The data were collected with 3 different sensors: Myo armband records the sEMG, DVS [3] records the so call DVS events, and DAVIS [4] collects DVS events and APS frames[1]. The choice to include both DVS and DAVIS in the dataset is due to the fact that DVS has lower resolution (128x128) and can be easily run on a mobile application for real-time inference. On the other hand, DAVIS offers the possibility to compare the performance of DVS events with APS frames. An example of the different kinds of recorded data is shown in Fig. 1. The Myo armband is composed of 8 equally spaced non-invasive sEMG channels that provide a sampling rate of 200Hz. The armband was placed approximately around the middle of the forearm. The DAVIS and DVS cameras were mounted on a 3D moving system, that was moved randomly to simulate the saccade movements. In this way, we can generate relative movement that can be detected by the DVS/DAVIS camera in a more biologically realistic approach and without introducing noise in the Myo sensor. The subjects were standing in front of the cameras setup with a white background to avoid having a dynamic scene. All the subjects were recorded with the same light conditions. The EMG recording was synchronized with the events/frames acquisition by restarting the camera zero-timestamp at every new session. The dataset contains recordings of 10 subjects. Each subject performed 3 sessions, where 5 hand gestures (pinky, elle, yo, index and thumb) were recorded 5 times, each lasting for 2s. Between the gestures, a relaxing phase of 1s is present where the muscles could go to the rest position, removing any residual muscular activation.

### C. Feature extraction

This section describes the steps of processing needed to obtain the features used for classification. The classification was carried out with two different classifiers, which calls for the need of extracting different kinds of features. The processing steps were kept to a minimum to limit overall system's latency during real-time operation.

[1]Zenodo link: https://zenodo.org/record/3228846#.XP5_cC-B3yx

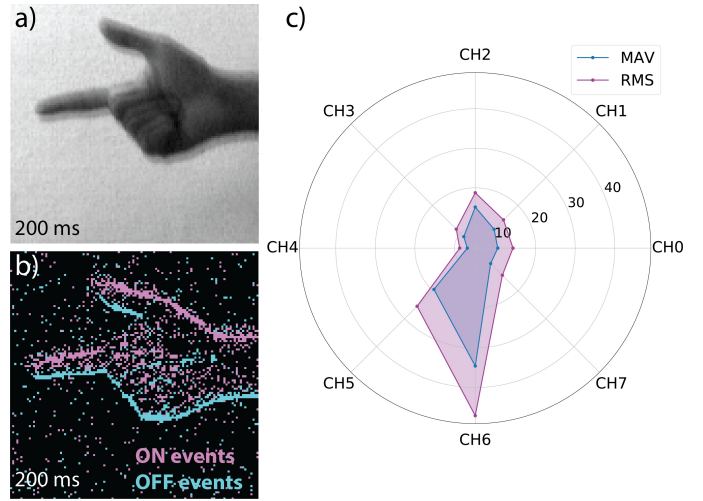

Fig. 1: Example of data from the dataset: a) APS frame (DAVIS), the image is blurred due to the long exposition (200ms); b) DVS frame (DAVIS), generated by accumulation of events; c) EMG features for the 8 channels of the Myo.

*1) EMG feature extraction:* For the EMG signal, we selected two time domain features traditionally used in the literature [7], namely MAV and RMS which are calculated over a certain window of length $T$ms.

We calculated the features for each channel separately and concatenated the resulting values in a vector $\mathbf{F}(n)$:

$$\mathbf{F}(n) = [F(x_1), ..., F(x_C)]^T \quad (1)$$

where $\mathbf{F}$ is MAV or RMS, $n$ is the index of the window and $C$ is the number of EMG channels. The final feature vector $\mathbf{E}(n)$ for window $n$ used for the classification is obtained by concatenating the 2 single feature vectors:

$$\mathbf{E}(n) = [\mathbf{MAV}(n)^T, \mathbf{RMS}(n)^T]^T \quad (2)$$

*2) Event frames feature extraction:* In order to use the DVS events for gesture classification with conventional algorithms, we need to turn the stream of events into frames, which we refer to as event frames. These frames are generated by accumulating the events occurring in a fixed time window of length $T$ ms. DVS frames can so be synchronized with the APS frame and EMG signal. In particular, we consider all the events within the time window (ignoring their polarity) and count how many events occur for each of the pixels separately. We then transform the event count frame into gray scale by min-max normalization. The event frames obtained from the DVS and the DAVIS sensors have a resolution of 128x128 and 180x240 pixels respectively. Since the region with the hand gestures does not fill the full frame, we extract a 60x60 pixels patch that allows us to significantly decrease the amount of computation needed during the visual feature extraction. In the case of the DAVIS, we extract a 120x120 patch and resample it to a 60x60 patch. This patch is extracted by detecting the hand in the frame with the zeroth order moment. This approach is reliable for event frames and has very low computational complexity. The patches are used as the input for the CNN while we extract other features for the Support Vector Machine (SVM). In particular, we extracted Histogram of Gradients (HOG) features [13] which have been extensively used for hand gesture recognition.

*3) Camera frames feature extraction:* The camera frames are obtained from the DAVIS sensor in gray scale and are averaged over the time window of length $T$ ms. In order to extract the hand gestures and the frame features, we follow a similar procedure to the one described in Section II-C2 for event frames. The hand detection is performed on the DAVIS event frames. Once the center of the hand gesture is detected, we extract a patch of 120x120 pixels from the APS frame, since the DAVIS resolution is much higher than that of the DVS. We then subsample the patch to a size of 60x60 in order to match feature size for both the event frames and the APS frames, and have fair comparisons. As in the case of event frames, the 60x60 patches are used by the CNN while the SVM uses HOG features extracted from these patches.

### D. Offline classification

For both the single sensors and the sensor fusion, we first trained and tested the models offline and then ported them into the mobile app. As stated above, we used two classifiers: an SVM and a CNN. We trained and tested the models on all the different modalities, namely EMG, DVS which are the frames calculated from the DVS events, DAV which are the frames obtained from the DAVIS events, FRM which are the APS frames of the DAVIS, FUS-DVS which is fusion of EMG and DVS, FUS-DAV which is fusion of EMG and DAV and finally FUS-FRM which is fusion of EMG and FRM.

*1) SVM classifier:* We trained an SVM classifier for each modality separately. In particular, classic features for the EMG signal and HOG features for the images. In case of the fusion, we concatenated the vectors obtained for the features of each of the modalities. We considered both a SVM classifier with a linear kernel and one with an Radial Basis Function (RBF) kernel. In the case of the RBF kernel, we selected the standard paramater $\gamma = 1/d$ where $d$ is the number of features, for all the modalities. The slack parameter of SVM was found by means of best average performance with 5-fold cross-validation for each modality separately.

*2) CNN classifier:* We trained several CNN architectures trying to minimize the size of the network while achieving the best accuracy. We finally chose the LeNet-5 architecture [14] for DVS and DAVIS data, and a slightly modified version with one-dimensional kernels and no pooling layers for EMG data. The fusion is then made with 5 Perceptrons (corresponding to the 5 classes of our dataset) that are fully connected to the two CNNs outputs. The training is done with TensorFlow using Adadelta gradient descent in two steps: first, we train the two uni-modal CNNs, then we train the Perceptrons layer based on the trained CNNs output activities.

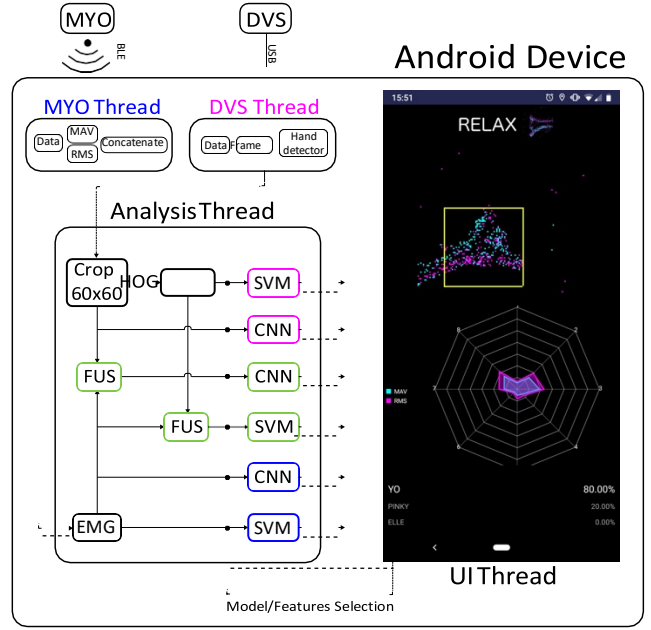

Fig. 2: Schematic of the Android application.

### E. Mobile Sensor Framework

The 'RELAX' application was developed via the standard Android Application Framework consisting of mainly Java routines and was run on a Pixel 3 mobile device. The connection to the DVS happens via USB using the libusb bindings already present in the Android OS. The communication protocol follows the one defined in jAER [15]. The connection to the Myo happens instead via Bluetooth.

Overall, the application flow, sketched in Figure2, consists of 4 different parallel threads. Two threads for the data collection, one processing thread and one UI thread. For the two data threads, one thread is used to pull the data from the DVS and construct DVS frames, while the other thread collects the data from the Myo and calculates the EMG features. The DVS thread communicates the frames both to the UI thread, so that they can be displayed and to the processing thread. The Myo data thread also communicates the features to the UI thread, to display them, and to the processing thread.

The processing thread reads the user defined parameters such as modality and model and does the needed processing via either the SVM or the CNN. Finally, the processing thread communicates to the UI thread the classification output to be displayed.

### III. RESULTS AND DISCUSSIONS

The data collected in the dataset described in Section II-B was used to train all the classifiers offline. The results are reported in Table I and show the classification accuracy for the different models and features. Even though we have tested all the models and features with window sizes $T \in \{100, 150, 200, 250\}$ ms, we only show results for the window size of $T = 200$ ms. As known from literature [16], EMG classification benefits from longer windows. We selected 200 ms since lower windows would yield worst results while a window of 250 ms did not significantly improve the results and require more computation for the feature extraction. As we can see, the results using EMG only are not satisfactory with a linear model but can be successfully improved with non-linear models. The difficulty resides on the training of a multi-subject model. Among the images, the APS frames have the best overall performance. Nevertheless, the DAVIS frames are not far behind. For the DVS frames, we can see that SVM yields arguably worse results than the other two image modalities. This is not true for the CNN which yields satisfactory and comparable results for all image modalities, including DVS frames. Overall, we can clearly see the advantage of the sensor fusion which achieves better results than the single modalities in all three fusion cases considered. Indeed, the best fusion accuracies using CNN improve by an average of $7.5\%$ compared to the best uni-modal accuracies. Hence, the CNN model with FUS-DVS is the best compromise in terms of accuracy and computational complexity.

TABLE I: Performance of hand gesture classification for different models. We report average and standard deviation over 5-fold cross-validation.

| Features | Frame (ms) | Accuracy % | | |
|---|---|---|---|---|
| | | SVM Linear | SVM RBF | CNN |
| EMG | 200 | 54.4 ± 1.0 | 76.9 ± 0.8 | 82.7 ± 0.8 |
| DVS | 200 | 74.5 ± 0.3 | 84.3 ± 0.7 | 90.0 ± 0.3 |
| DAV | 200 | 79.2 ± 1.9 | 85.4 ± 1.3 | 91.2 ± 1.3 |
| FRM | 200 | 81.1 ± 0.9 | 88.4 ± 1.6 | 91.9 ± 0.7 |
| FUS-DVS | 200 | 82.6 ± 0.5 | 88.7 ± 0.8 | 98.3 ± 3.4 |
| FUS-DAV | 200 | 84.5 ± 0.4 | 89.3 ± 0.5 | 98.5 ± 3.0 |
| FUS-FRM | 200 | 86.2 ± 0.7 | 93.4 ± 0.7 | 98.8 ± 2.3 |

Given the offline results, we deployed and tested the CNN models on the smartphone. These online tests were performed on two subjects, for a total of 200 hand gestures, one of which was not part of the training dataset. Moreover, the tests were carried out in a room with different lighting conditions than the ones in which the dataset was recorded. The real-time classification yields $72\%$ and $74\%$ for the single EMG and DVS sensors respectively. Using the fusion CNN, the classification accuracy reaches $85\%$, showing the benefits of fusion in real-time classification as well. On the smartphone, the fusion CNN has an inference time of about $18ms$, a CPU consumption of $24\%$ and a memory consumption of $392MB$ of which only $144MB$ are used by the CNN.

## IV. CONCLUSIONS

In this work, we presented a system that allows to recognize static hand gestures using a smartphone computational capabilities. The classifier fuses the complementary signals of EMG and images (event and camera frames) to improve the classification accuracy. The mobile application can be useful for close-loop systems in calibration and prosthetic control for personalize medicine. To demonstrate the systems's capability, we recorded a new dataset for offline training where we confronted SVM to CNN, then we performed online test achieving $85\%$ of gesture recognition accuracy.

ACKNOWLEDGMENT

The authors would like to acknowledge the 2019 Capocaccia Neuromorphic Workshop and all its participants for the fruitful discussions. This work is supported by the EU's H2020 MSC-IF grant NEPSpiNN (Grant No. 753470), SWITCHBOARD ETN (Grant No. 674901), NeuroTech project, the SNSF grant No. 200021_172553 and Toshiba Corporation. Finally, we thank Prof. B. Miramond, Prof. S. Liu, Prof. T. Delbruck and Prof. G. Indiveri.